\documentclass{article}

\usepackage{PRIMEarxiv}

\usepackage[utf8]{inputenc} 
\usepackage[T1]{fontenc}    
\usepackage{hyperref}       
\usepackage{url}           
\usepackage{booktabs}       
\usepackage{amsfonts}       
\usepackage{nicefrac}      
\usepackage{microtype}      
\usepackage[table,xcdraw]{xcolor}
\usepackage{subcaption}
\usepackage{pgfplots}
\usepackage{colortbl}
\usepackage{float}
\usepackage{lipsum}
\usepackage{fancyhdr}       
\usepackage{graphicx}       
\graphicspath{{media/}}

\title{Neuromorphic Circuit Simulation with Memristors: Design and Evaluation Using MemTorch for MNIST and CIFAR}
\author{Julio Souto Garnelo; Guillermo Botella; Daniel García; Raúl Murillo; Alberto del Barrio \\
Dpto. Arquitectura de Computadores y Automática. Universidad Complutense de Madrid \\
\texttt{\{jusouto, gbotella, daniel10, ramuri01, abarriog\}@ucm.es}
}

\begin{document}
\maketitle

\begin{abstract}
Memristors offer significant advantages as in-memory computing devices due to their non-volatility, low power consumption, and history-dependent conductivity. These attributes are particularly valuable in the realm of neuromorphic circuits for neural networks, which currently face limitations imposed by the Von Neumann architecture and high energy demands. This study evaluates the feasibility of using memristors for in-memory processing by constructing and training three digital convolutional neural networks with the datasets MNIST, CIFAR10 and CIFAR100. Subsequent conversion of these networks into memristive systems was performed using Memtorch. The simulations, conducted under ideal conditions, revealed minimal precision losses of nearly 1\% during inference. Additionally, the study analyzed the impact of tile size and memristor-specific non-idealities on performance, highlighting the practical implications of integrating memristors in neuromorphic computing systems. This exploration into memristive neural network applications underscores the potential of Memtorch in advancing neuromorphic architectures.
\end{abstract}


\section{Theoretical introduction}
\subsection{Von Neumann bottleneck}

\begin{figure}[h!]
\centering
\includegraphics[width=0.35\linewidth]{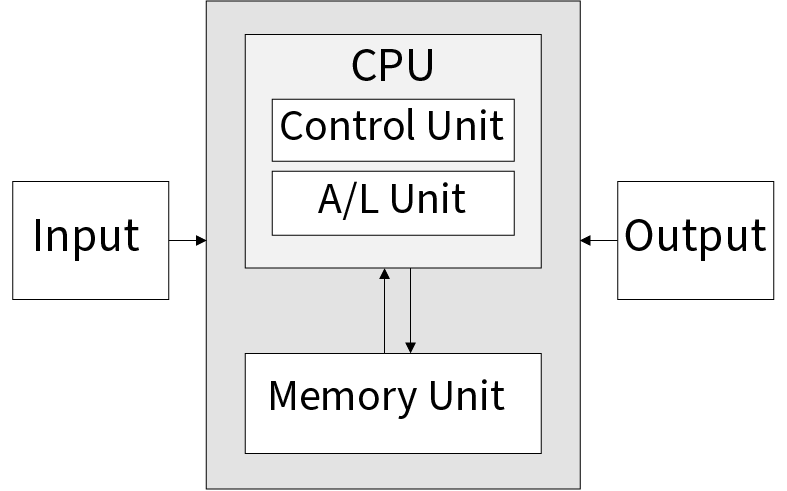}
\caption{Basic Layout of Von Neumann Architecture.}
\label{fig:test}
\end{figure}

In the present day, the majority of computers use or are based on Von Neumann’s architecture, first proposed by John Von Neumann in 1945. This architecture stores both instructions and data in the same memory system, treating them as the same type of information, which allows for greater flexibility as new instructions can be entered into memory. The design features a Central Processing Unit (CPU), comprising a Control Unit and an Arithmetic Logic Unit, alongside an Input/Output subsystem and a unified memory for both data and instructions. The interpretation of data and instructions depends on their context  \cite{VN}. Although CPU speeds have significantly outpaced those of memory access, a major imbalance in cycle times persists, limiting overall performance. This limitation, known as the Von Neumann bottleneck, is primarily due to delays in memory access that can cause the CPU to remain idle. Additionally, the latency of the system bus connecting these components is a critical determinant of the total processing speed.

\subsubsection{PIM: Proccessing in Memory}
There have been many strategies utilized to overcome the aforementioned bottleneck. One approach has been to modify the architecture to minimize data movement by conducting the processing in memory. This implementation in the architecture allows memory-intensive devices to overcome the so-called ``memory wall". We can distinguish three different paths to implement this methodology \cite{VN1}. As an architectural issue, focusing on combining memory chips and processing units, for instance by constructing 3D arrays. Secondly, utilizing peripheral circuits within memory sub-arrays as a manner of conducting the processing in memory. And lastly, and perhaps the most relevant to our case of study, by manufacturing devices that allow in-situ computation in memory. For example, Resistive Random Access Memory (RRAM) devices. This specific approach makes use of Kirchoff's laws and Ohm's law in order to carry out matrix calculations.
\subsection{Neuromorphic circuits}
Following the discussion on processing in memory devices, the interest on diminishing the amount of energy of today's computational tasks has added a special relevance to the construction of circuits taking inspiration on the brain, as it is very energy-efficient. As a result, the notion of neuromorphic circuits appeared by trying of reproduce the functioning of neurons and synapses in the brain \cite{Neuro}.
\\ 
Our brain's neurons transmit signals that vary according to the strength of the synaptic connection. This strength is updated during learning. Similarly, a weight is assigned to a physical characteristic of, for instance, a resistive device, such as a conductive state. The weight must be a parameter that can be stored and be able to change in the artificial learning process.\\
Several devices are being studied for their applicability in neuromorphic circuits: spintronics, which use the electron's spin to carry out computations; memristors, devices that offer a resistance that remembers its current history; synaptic transistors, hardware components -less costly in terms of energy than the memristor- that remember their voltage history and have, until recently, been devices limited to impractically low operating temperatures. A few more candidates exist, but we will focus on the memristor. It is worth to comment briefly the recent advances of the synaptic transistors, a device that utilizes a moiré pattern that presents ferromagnetic properties at room temperature \cite{syntrans}, it mimics a neuron as it displays a moiré potential which captures electrons in the same way synaptic vesicles accumulate neurotransmitters. Nowadays, the first place as the the optimal device for neuromorphic circuits is very disputed, as each proposed device offers different improvements and limitations.\\
Building on the innovations in Processing in Memory (PIM) and the significance of minimizing data movement to overcome the 'memory wall,' it becomes crucial to explore effective architectural modifications. These modifications not only address the limitations of traditional computing architectures but also pave the way for advanced computational technologies. Integrating processing capabilities directly within memory arrays reduces latency and energy consumption associated with data transfer, leading to a resurgence in analog computing. Field-Programmable Analog Arrays (FPAAs) serve as key examples of how processing capabilities can be smoothly integrated within memory arrays \cite{clusterfpaas}.
\subsection{Neural Networks}
Neural Networks (NNs) may have many layers with thousands of artificial neurons. Figure 2 shows a typical NN structure composed of input layers, hidden layers, and output layers. As mentioned earlier, neurons communicate via synapses. This behavior is replicated by artificial neurons, substituting the synapse by an input and the strength by the weight; then, activation is determined by a mathematical function.
\begin{equation}
    y=\phi(\Sigma_jx_jw_j-b)
\end{equation}
Here, the b corresponds to the bias, an inner parameter of the artificial neuron that participates in the linear relationship to displace the outcome. The $\phi$ is the chosen activation function.
\begin{figure}[h!]
\centering
\includegraphics[width=0.45\linewidth]{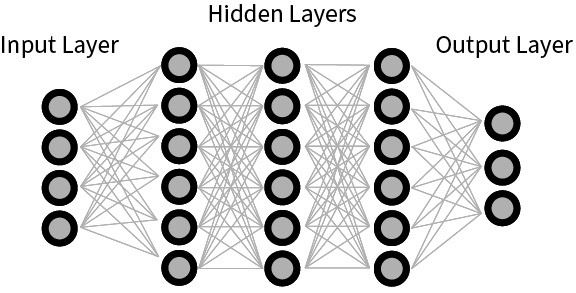}
\caption{Architecture of a Multi-Layer Perceptron (MLP).}
\label{fig:test}
\end{figure}
As the weights dictate whether signals are propagated or inhibited, having a correct value is crucial. Thus, learning algorithms are used to tune the values of the weights based on the error of the network in a process called training, during which patterns are learned in order to make predictions. This process is called backpropagation. We can compute the error of each neuron by comparing the outcome to the desired output:
\begin{equation}
    E_i(x,w,d)=O_i(x,w)-d_i
\end{equation}

The total error of the neural network is the sum of all individual errors. Using the gradient descent method, the network updates its weights based on their influence on the error, according to next formula:

\begin{equation}
    \Delta w_{ij}=-\eta \frac{\partial E }{\partial w_{ij}}
\end{equation}
There are several parameters that are set beforehand, called the hyperparameters. For instance, the learning rate of the network, which is the pace at which the algorithm updates its values during training. Other examples are batches, the number of training samples passed through at once during one training step; epochs, the number of learning  repetitions on the same dataset; or the amount of hidden layers.
\subsubsection{Convolutional Neural Networks}
Convolutional Neural Networks (CNNs) are a class of neural networks that have become the standardized network for computer vision tasks.\\ 
These functionalities benefit greatly from the ability of the CNNs to extract features from an input image. The key components are convolutional layers, which create a feature map from an input image and later send it to an activation function. In a way, convolutional layers simplify images by keeping their most recognizable features in order to classify them. In the interest of understanding CNNs, we can use a common model such as VGG, represented in Figure 3, to explain the different layers \cite{CNN}:
\begin{figure}[h!]
\centering
\includegraphics[width=0.55\linewidth]{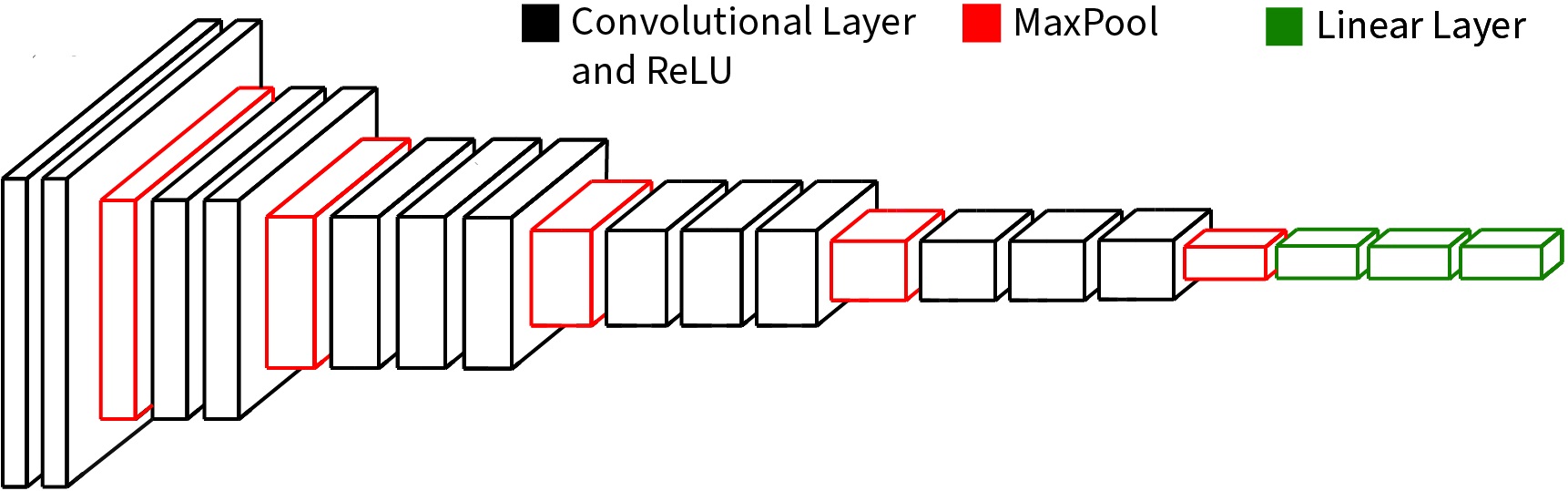}
\caption{Architecture of a VGG-style CNN.}
\label{fig:test}
\end{figure}
\begin{itemize}
    \item Convolutional layer: 
    This layer scans the entire input image using sections known as kernels, which extract patterns through an element-wise dot product between the input and the kernel weights. After this multiplication, a bias term is added. For RGB images, the outputs from the three color channels are combined element-wise before adding the bias. The stride specifies the movement of the kernel across the image after each operation. The dimensions of the output image change according to the following formula, which accounts for the sizes of each variable:
    \begin{equation}
        Output=\frac{Input+2\cdot Padding-Kernel}{Stride}+1
    \end{equation}
    Padding is used to maintain a certain image size, which can be useful at times to extract more features and prevent early downsizing.
    \item ReLU layer: Rectified Linear Unit is a type of activation function that is very commonly used in NNs. The output of the convolutional layer is passed element-wise through the activation function:
    \begin{equation}
        ReLU(x)=max(0,x)
    \end{equation}
    This functions adds non-linearity to the CNN which is needed to prevent having an output that corresponds to a linear combination of the inputs. Non-linear decision boundaries are a necessity in order to learn complex images.
    \item Pooling layer: Similarly to a convolutional layer, a Kernel and a Stride have to be selected in the pooling layer. This function sweeps the feature map created by the convolutional layer and takes the maximum value of each Kernel, which effectively decreases image size.
    \item  Fully-connected layer: It is composed of a flatten function that eliminates the area dimension of the image for it to be used as an input for the linear layers. The primary function of these linear layers is to classify the image based on the features extracted by previous layers in the network, effectively mapping the learned features to specific outputs.
\end{itemize}

Several techniques can be employed to improve the accuracy and stability of convolutional neural networks (CNNs). For instance, batch normalization standardizes the inputs to each layer within a network, facilitating smoother and more stable training \cite{Batch}. Additionally, dropout layers enhance model generalization by randomly omitting a proportion of the units during training, thereby preventing the units from co-adapting too closely. When configured with an optimal dropout rate, this technique helps mitigate overfitting.

\subsection{Memristors}

Circling back to devices for processing in memory, we are focusing on memristors. Since Leon O. Chua \cite{OGchua}, an electrical engineer, published 'Memristor - The Missing Circuit Element' in 1971 [2], this device has been regarded as the missing basic circuit element. Chua postulated its existence, noting that until then, there were three primary components in a circuit, each corresponding to two fundamental circuit variables: magnetic flux, voltage, charge, and current. Resistors relate current and voltage; inductors link magnetic flux and current; and capacitors connect charge and voltage. The only missing device was one that could establish a relationship between charge and magnetic flux.

It was not until 2008 that a research group at Hewlett-Packard Labs claimed to have discovered this missing element \cite{found}. Controversy arose because the device was not entirely new and did not establish the relationship between memory and magnetism as initially theorized. In fact, this new device could operate without magnetism. Nevertheless, Leon O. Chua recognized the discovery. In 2014, he published 'If it’s pinched it’s a memristor,' a paper in which he broadened the definition of memristors \cite{ifChua}. He emphasized that if a semiconductive device exhibited a pinched hysteresis loop, then it was a memristor, even if it was not ideal.

\subsubsection{Physical characteristics}
Memristors are characterized by a simple structure of metal-insulator-metal. It can be described as a sandwich where the two outside layers are formed by electrodes and, in the middle, there can either be a insulator or semiconductor depending on the device.\\
As a general explanation of the functioning of the device it will be useful to analyze HP's memristor, which offered a bilayer of Pt electrodes and a middle layer of TiO$_2$ \cite{found}. The TiO$_2$ was doped, creating a zone of oxygen vacancies and a zone without them. As a consequence, a boundary is created in the resistive layer between the two zones. As the oxygen-starved TiO$_2$ offers electrons to carry the current, these vacancies reduce resistance. Nevertheless, it is necessary to take into account that vacancies have positive charge, which consequently causes the boundary between doped and undoped zones to move, effectively altering resistance. \\
Depending on the polarity of the current, the boundary moves towards either side, setting the memristor to a high resistance state -or HRS-, if the doped region is narrow, or low resistance state -or LRS-, if the doped region is wide \cite{MEM}. This is determined by the state variable w \cite{wmem} and can be observed in Figure 5.\\
In addition, memristors do not store any current, thus, the set resistance state is remembered and depends on current history. This makes the memristor a candidate for neuromorphic devices, since it displays the plasticity of neurons.
\begin{figure}[h!]
    \centering
    \includegraphics[width=0.40\linewidth]{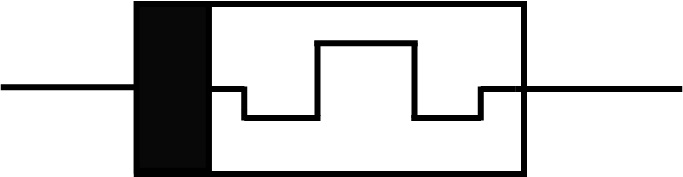}
    \caption{Memristor Circuit Symbol.}
    \label{fig:litio47}
\end{figure} \\
The internal materials may vary, as there are different types of memristors. For example, the HP memristor falls under the anion memristors \cite{wmem}. Meanwhile, there are also cation memristors and dual-type ones, which mix the migration of both anions and cations whithin the resistive layer. In order to explain their properties, a brief comment on possible materials for each component will be made.
\begin{figure}[h!]
    \centering
    \includegraphics[width=0.50\linewidth]{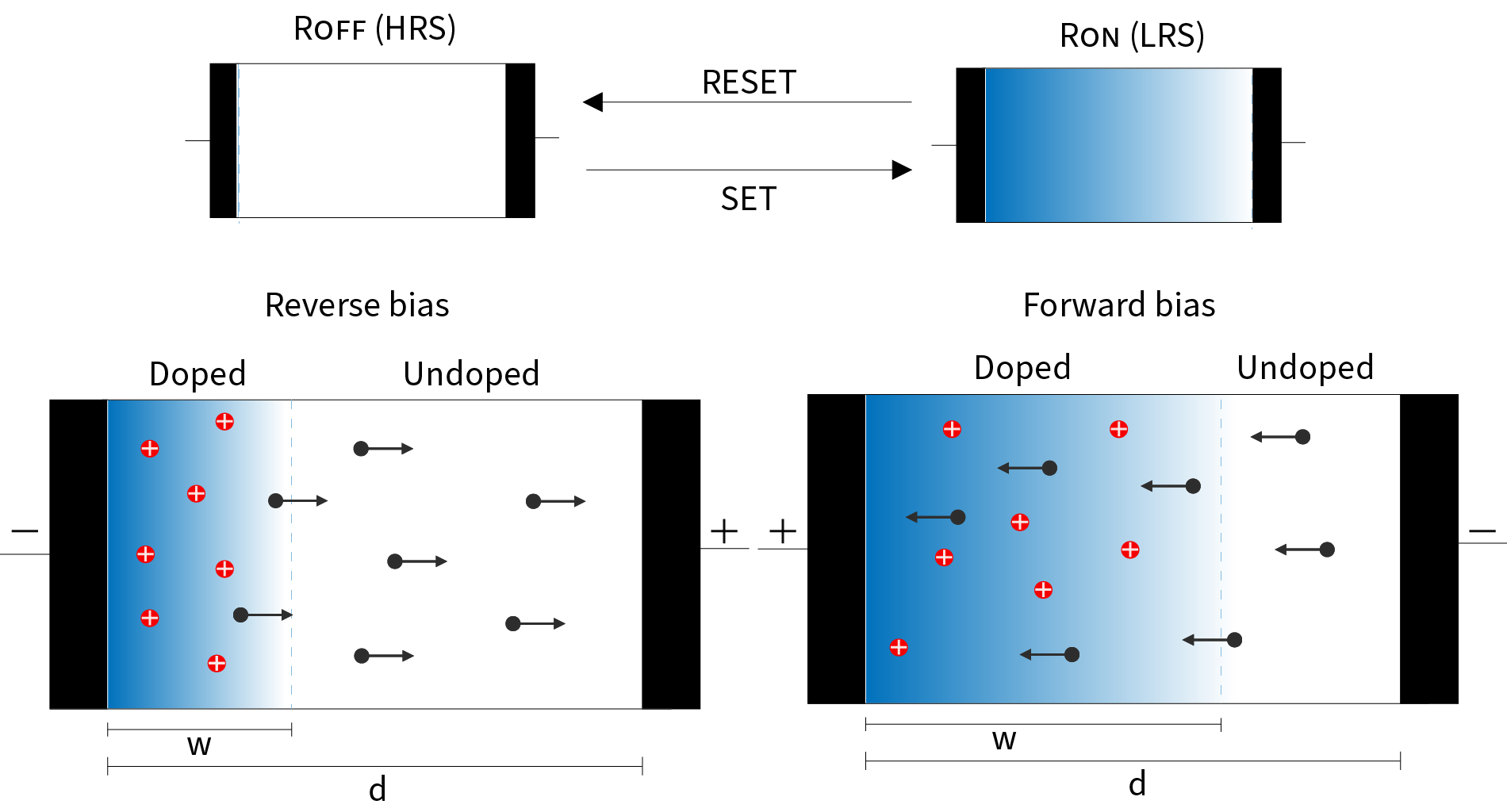}
    \caption{Memristor states and switching mechanism.}
    \label{fig:litio47}
\end{figure}
The electrodes may play a role in the resistive switching present in memristors or solely let current through. For instance, in the case of cation memristors, electrochemically active metals are used, mainly Cu and Ag. This creates conductive filaments, narrow pathways that offer a route for ions to migrate which form thanks to electrochemical reactions in the electrode. There are other materials that do not alter the behavior noticeably, commonly inert metals such as Pt and Au. Additionally, other materials like graphene or several alloys can be used depending on the technical requirements of the device. \\
The key component is the resistive layer, as it ultimately determines the behavior of the memristor. It can be composed of inorganic or inorganic materials \cite{typesmem}. The former offers more stability, while the latter is less costly production-wise. \\
Inorganic materials that can be used in a memristive device may be binary oxides, perovskites and 2D materials. Binary oxides, as TiO$_2$, allow for high stability, long durability, fast switching speeds and compatibility with Complementary Metal-Oxide Semiconductor processes for fabrication. Perovskites, which are compounds that share the structure of CaTiO$_3$, possess optoelectronic properties that make them optimal for memristors, yet sustainability is an issue as well as its production, as they are not compatible with the standardized CMOS process. Inorganic 2D materials can be used to produce memristors that display low-power consumption and flexibility but little scalability. 
\subsubsection{Properties}
Memristors stand out by its low power consumption, perhaps one of their most relevant qualities nowadays, considering the amount of energy consumed in machine learning. Analog computing in memory with memristors tackles both the need to surpass the memory wall as well as the energy issue.\\
Further, memristors are two-terminal devices that present non-volatility, the ability to store data without the necessity of current going through. This property allows for the construction of ReRAM devices or Resistive Random-Access Memory \cite{RRAM}. These devices support the creation of densely packed computational circuits and can endure a great amount of cycles.\\
Memristors can switch states very quickly, which is very useful in high-speed data processing. This switching behavior can be seen in Figure 6.a, a current-voltage graph. Current is plotted logarithmically, which accentuates the leap from low currents nearing 0 voltage to higher values of current. We can observe how the current nearly plateaus at high values. The mechanism explained in the previous section is clearly seen: increasing voltage increases current in a non-linear manner, and the same can be said for decreasing negative voltages as the memristor is bipolar. Both set and reset paths are observed on each side of the RS graph.
\begin{figure}[h!]
  \centering
  \begin{subfigure}[b]{0.47\textwidth}
    \centering
    \includegraphics[width=\textwidth]{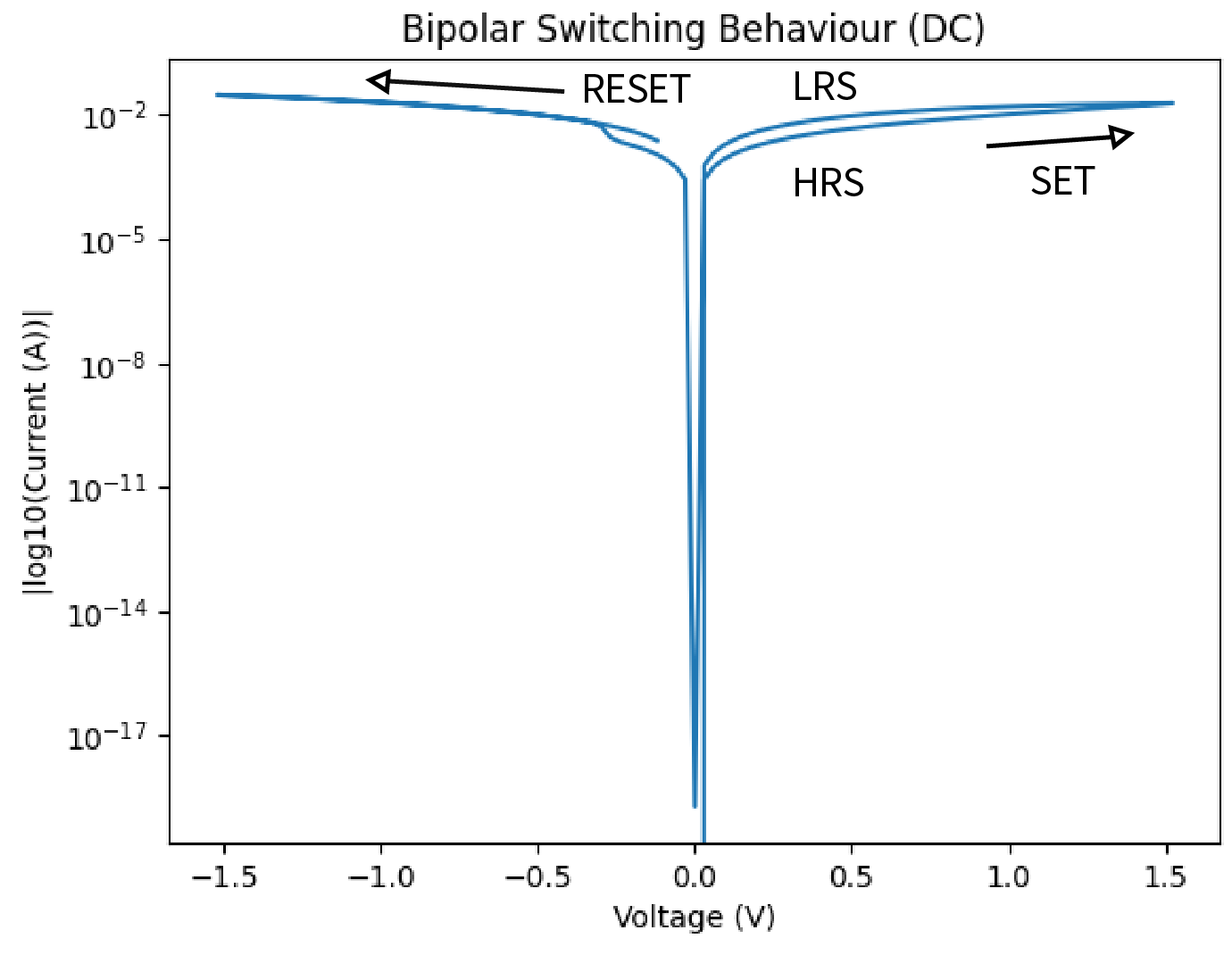}
    \caption{Switching Behavior.}
    \label{fig:first}
  \end{subfigure}
  \hspace{0.01\textwidth}
  \begin{subfigure}[b]{0.48\textwidth}
    \centering
    \includegraphics[width=\textwidth]{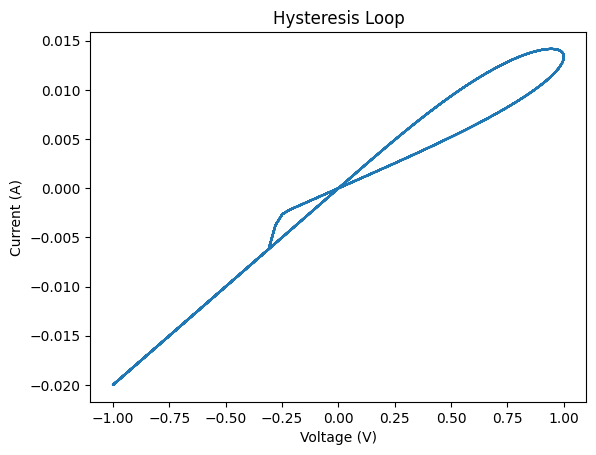}
    \caption{I-V Hysteresis loop.}
    \label{fig:second}
  \end{subfigure}
  \caption{(a) Bipolar switching behavior and (b) hysteresis loop simulated with Memtorch, a simulation framework for memristive crossbar arrays using VTEAM memristor model. Source: Own Memtorch simulation.}
  \label{fig:graphs}
\end{figure} \\
The hysteresis loop is the representation of the ferromagnetic behavior of memristors, showing the dependence of a system on its history, usually in terms of magnetic flux (B) and magnetic field strength. In Figure 6.b it is represented in the I-V (current-voltage), meaning that current can be different for the same voltage depending on history. It crosses itself at (0,0), what we typically refer to as a pinched loop. Thus, when there is no voltage, the current is 0 as well. Nevertheless, the resistance state is remembered. Depending on the frequency of the current, the hysteresis loop can appear closer to a line if the frequency is high, or wider if the frequency is lower \cite{MEM}.\\
Commercial Ag/Ge2Se3/SnSe/Ge2Se3 memristors have been characterized experimentally in the literature\cite{memcha} through I-V plots derived from voltage-ramped signals. Variability in the devices was assessed using automated extraction methods. Both the I-V behavior and variability were successfully modeled using the Dynamic Memdiode Model, with model parameters refined through a genetic algorithm.
\subsubsection{Memristor array}
Delving further into memristive analog computation requires explaining the arrangement of memristor crossbar arrays. This architecture allows processing in memory by carrying out matrix-vector multiplications, which are essential for CNNs. \\ 
Non-volatile memory elements as memristors allow us to encode numerical values to perform calculations as analog conductance states. Thus, we exploit the memoristic ability of a device to store information used to perform Matrix Vector Multiplications by means of Kirchoff's and Ohm's laws. Conductance acts as a counterpart to resistance, measuring the memristor's ease to let electricity flow.
\begin{equation}
    G=\frac{I}{V}=\frac{1}{R}
\end{equation}
Conductance depends on the state variable w of the memristor. Thus its behavior can be explained as a function of w:
\begin{equation}
    I(t)=G(w(t))\cdot V(t)
\end{equation}
\begin{equation}
    \frac{d}{d t}w(t)=F(w(t),V(t))
\end{equation}
Equation (8) showcases how the state variable changes in time depending on the voltage going through and the state of the memristor. Thus, the inherent non-linearity is characterized by these expressions \cite{designsim}.\\
\begin{figure}[h!]
    \centering
    \includegraphics[width=0.20\linewidth]{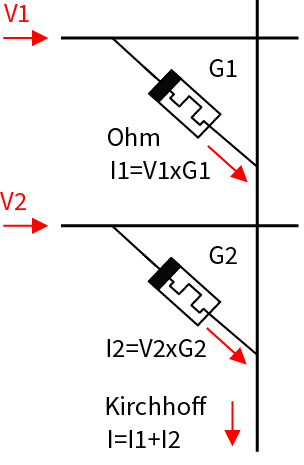}
    \caption{Element-wise Operations (partial Matrix multiplication) in a Memristor Crossbar Array.}
    \label{fig:litio47}
\end{figure}\\
The array represents a matrix of conductances corresponding to weights. It features a structure formed by Word Lines (WL), which connect rows of memristors originating from a Digital to Analog Converter (DAC), and Bit Lines (BL), which run perpendicular to the Word Lines and connect to an Analog to Digital Converter (ADC). The Word Lines input a vector that is encoded as voltages by the DAC. These voltages are then modulated by the conductance states of the memristors in the array \cite{sims}. In this way, the matrix multiplication is carried out analogically as represented in Figure 7 by means of circuit laws and elements. Finally, each Bit Line outputs a current that is a result of the conductance state of each memristor in the column and the input voltages (9). This device is commonly called a memristor-based accelerator as it improves the speed of computation.
\begin{equation}
    I_j=\Sigma_i V_i G_{ij}
\end{equation}
Loading the weight matrices into the array is a critical yet time-consuming task. When handling large matrices, multiple crossbar tiles are employed to prevent performance degradation and ensure accurate layer representation. To facilitate this, a best-fit algorithm is utilized. This algorithm seeks to allocate matrices to the smallest suitable tile. If a matrix cannot be accommodated within any available tile, it is divided and distributed across multiple tiles to ensure proper mapping.

As memristors do not have negative conductance states, for each weight matrix two crossbar arrays may be needed to account for negative weights \cite{memT2}. As a result, weights are separated into negative and positive values, each into a different array. When a MVM (Matrix-Vector Multiplication) is conducted, both are taken into account:
\begin{equation}
    AB=K\Sigma A[i,:](g_{pos}[i,j]-g_{neg}[i,j])
\end{equation}
Where the $B$ matrix is the one that we map and the $A$ matrix is the input that enters through the WLs. The value of $K$ has to be determined for each crossbar by performing a linear regression to the output current of each column and its corresponding output. The conductances are represented by g. Furthermore, weights must be translated into conductance values, which can be done by means of:
\begin{equation}
    g[i,j]=\frac{(g_{ON}-g_{OFF})(\sigma (w)[i,j]-w_{min})}{|w|_{max}-w_{min}}
\end{equation}
Firstly, $g_{ON}$ and $g_{OFF}$ represent the conductances corresponding to the high and low resistance states of the memristor, respectively. Then, $w_{min}$ and $w_{max}$ are the lowest and highest weights to map. We are considering the physical bounds, the resistance states, and the bounds in our matrices' weights. As a result, the weights are adjusted to the analog bounds of the device. Finally, $\sigma (w)$ corresponds to the weight being mapped.
\\
In certain crossbar arrays, there are transistors integrated next to each memristor as seen in Figure 8, allowing for the selection of individual devices. This proves useful when mapping matrices, a process implemented by applying voltages to the memristors.
Thus, a 1T1R (one transistor-one resistor) array enables individual mapping of each memristor without influencing adjacent ones, whereas a conventional 1R (one resistor) array requires corrective methods to manage influences on nearby memristors.
\begin{figure}[h!]
    \centering
    \includegraphics[width=0.45\linewidth]{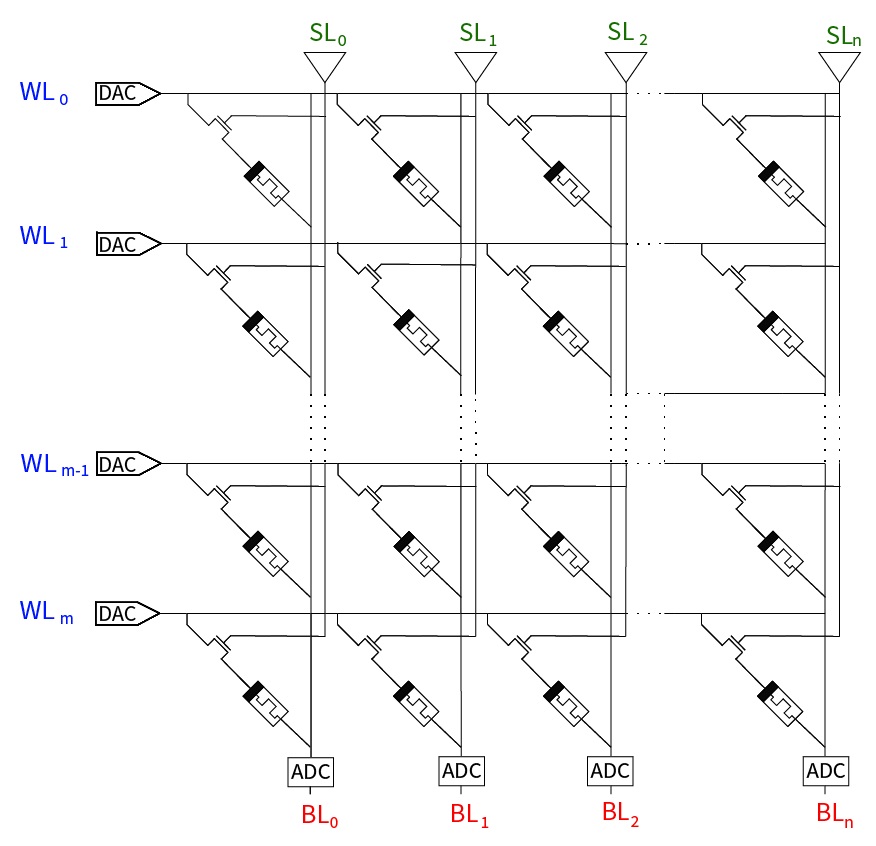}
    \caption{1T1R memristor crossbar array.}
    \label{fig:litio47}
\end{figure}
\subsubsection{Simulation Frameworks}
Regarding the current interest in memristor-based accelerators for their performance and low energetic consumption, there is a need for a reliable and multifaceted simulation framework. Nowadays, each simulator offers several great and unique qualities, but a general framework is lacking.\\
Most simulation frameworks focus on machine learning tasks. For instance, MNSIM was one of the first behavior-level simulators created for neuromorphic computing, enabling the customization of the design and estimating accuracies of memristive structures when computing. This model offers estimations of area, power consumption and latency, focusing on hardware design \cite{MNSIM}. NeuroSim also grants the tools to study the latency, area and power aspects of MBAs \cite{NeuroSIM}. It is built in a hierarchical multilevel manner that touches circuit, device and neural network topology, making it a robust framework to research device design. Lastly, PUMA assesses performance using an Instruction Set Architecture that allows for complex programming in conjunction with the efficiency of analog computing \cite{puma}.
\\
There are other frameworks that focus on non-ideal behavior, which is very relevant when simulating memristive-based devices, since analog computing presents errors due to electrical properties. DL-RSIM, for example, calculates the error rates of MBAs and takes them into account in the inference phase \cite{DLRSIM}. RxNN also considers non-idealities, such as parasitic behavior and process variations. It is centered around large-scale DNNs and it has shown proof that only modelling ideal behaviors offers poor results, as non-idealities cause important downgrades in accuracy \cite{rxnn}. GraphRSim analyzes the impact of these non-idealities in graph algorithms that use ReRAM devices as matrix multiplicators \cite{graphrsim}. IBM's Analog Hardware Acceleration Kit or aihwkit, on the other hand, focuses on optimization of NN algorithms for training and inference with analog chips \cite{awhkit}. The structure of a NN is very relevant when mapping weights on memristor arrays and thus an optimal model can greatly improve performance. \\ 
Simulations in this study will be conducted via Memtorch, a framework specifically designed to address non-ideal behaviors during inference \cite{memT1} that excels in replicating the nuanced behaviors of memristive systems under real-world conditions. It offers a detailed examination of non-linear and stochastic effects that are often overlooked by other frameworks.

\subsubsection{Memtorch}
Memtorch is an open-source framework based in Python that is compatible with the PyTorch MLlibrary focused on the modelling of non-ideal characteristics to study losses in accuracy \cite{memT2}.
\\
In order to utilize the simulator, we first need to construct and train a PyTorch model. Then it can be patched into a memristive one, writing weights into memristor crossbar arrays that Memtorch simulates, transforming the computation framework from digital to analog. This framework offers the possibility of taking into account non-idealities making it fairly useful in comparison tasks and assessment of precision losses. It is possible to account for finite conductance states, device faults, non-linearity, endurance and retention.

\section{Methodology}
\label{sec:headings}
Our objective is to asses the versatility and usefulness of Memtorch as a framework for conducting simulations of the matrix multiplications that comprise inference using memristor crossbar arrays. Initially, the purpose is to build the CNNs models to later convert to memristive models following the methodology displayed in Figure 9. Three models will be constructed, varying in sizes, each for a different dataset.
\begin{figure}[h!]
    \centering
    \includegraphics[width=0.60\linewidth]{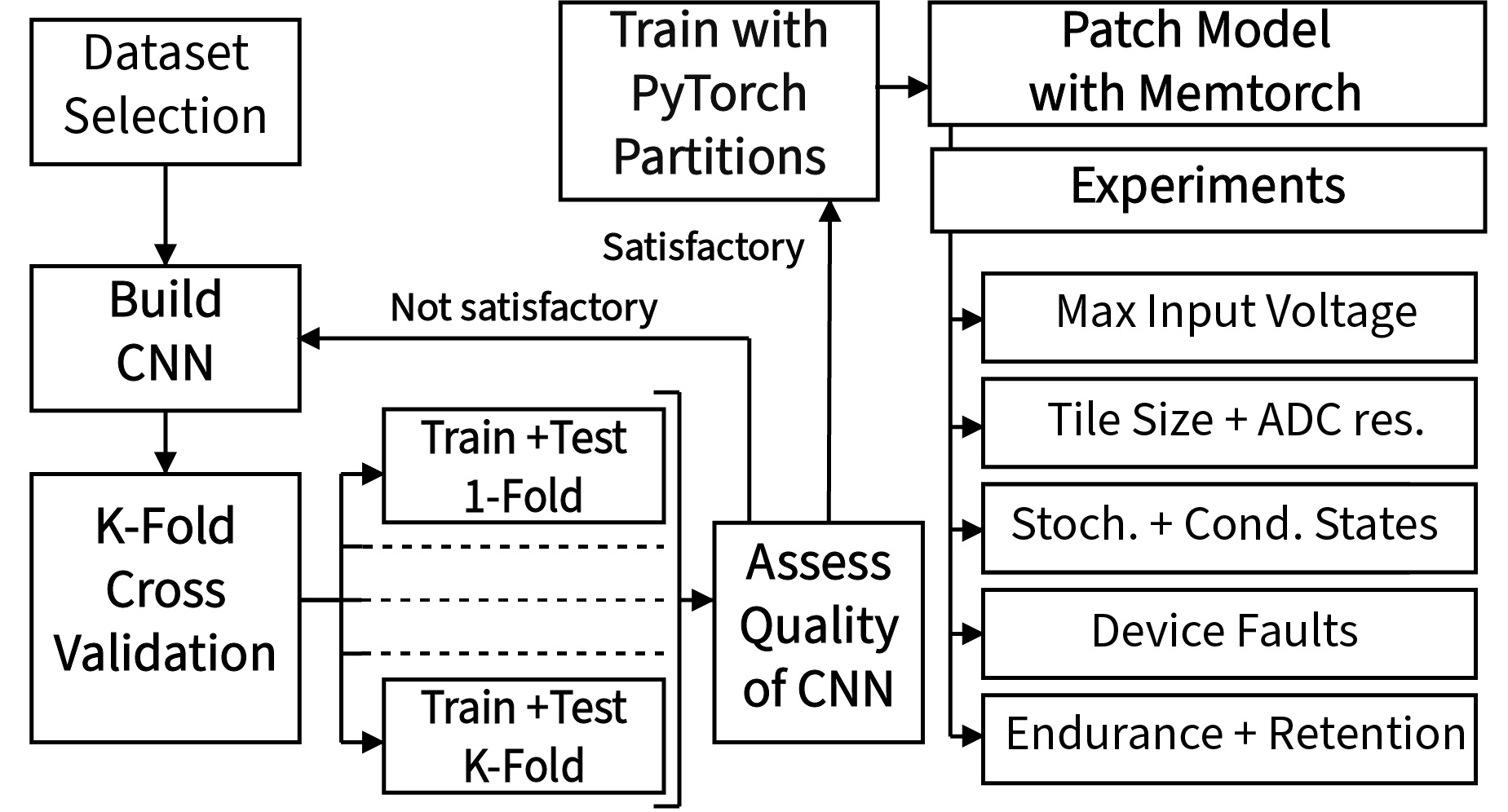}    \caption{Proposed methodology.}
    \label{fig:litio47}
\end{figure}
\subsection{Dataset selection}
As Pytorch offers several datasets that are readied for machine learning tasks we can use them to train our models. Each model built will be tailored to a chosen dataset from Table 1.
\begin{table}[h!]
\centering
\begin{tabular}{|c|c|c|c|c|c|}
\hline
\rowcolor[HTML]{EFEFEF} 
\multicolumn{1}{|c|}{\cellcolor[HTML]{EFEFEF}\textbf{Dataset}} & \textbf{Type} & \textbf{Size Images} & \textbf{Classes} & \textbf{Train Images} & \textbf{Test Images} \\ \hline
MNIST                                                          & BW            & $28\times28$               & 10               & 60,000                 & 10,000                \\ \hline
CIFAR10                                                        & RGB           & $32\times32$                & 10               & 50,000                 & 10,000                \\ \hline
CIFAR100                                                       & RBG           & $32\times32$                & 100              & 50,000                 & 10,000                \\ \hline
\end{tabular}
\caption{Datasets utilized and their characteristics.}
\end{table}
MNIST is commonly used for the training of toy models and it posed a great starting block for studying Memtorch's capabilities. It can also serve as tuning dataset as its simpler model can be patched faster, being useful for try and error processes of hyperparameter tuning. Nevertheless, the most relevant feature of the dataset is the simplicity of the images, which depict letters and numbers in gray scale such as that in Figure 10.a. This simplicity translates into easy-to-learn patterns that contribute to having robust and stable models.
\begin{figure}[H]
  \centering
  \begin{subfigure}[b]{0.228\textwidth}
    \centering
    \includegraphics[width=\textwidth]{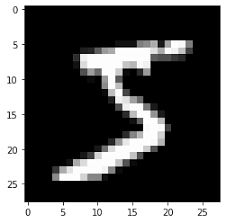}
    \caption{MNIST.}
    \label{fig:first}
  \end{subfigure}
  \hspace{0.01\textwidth}
  \begin{subfigure}[b]{0.20\textwidth}
    \centering
    \includegraphics[width=\textwidth]{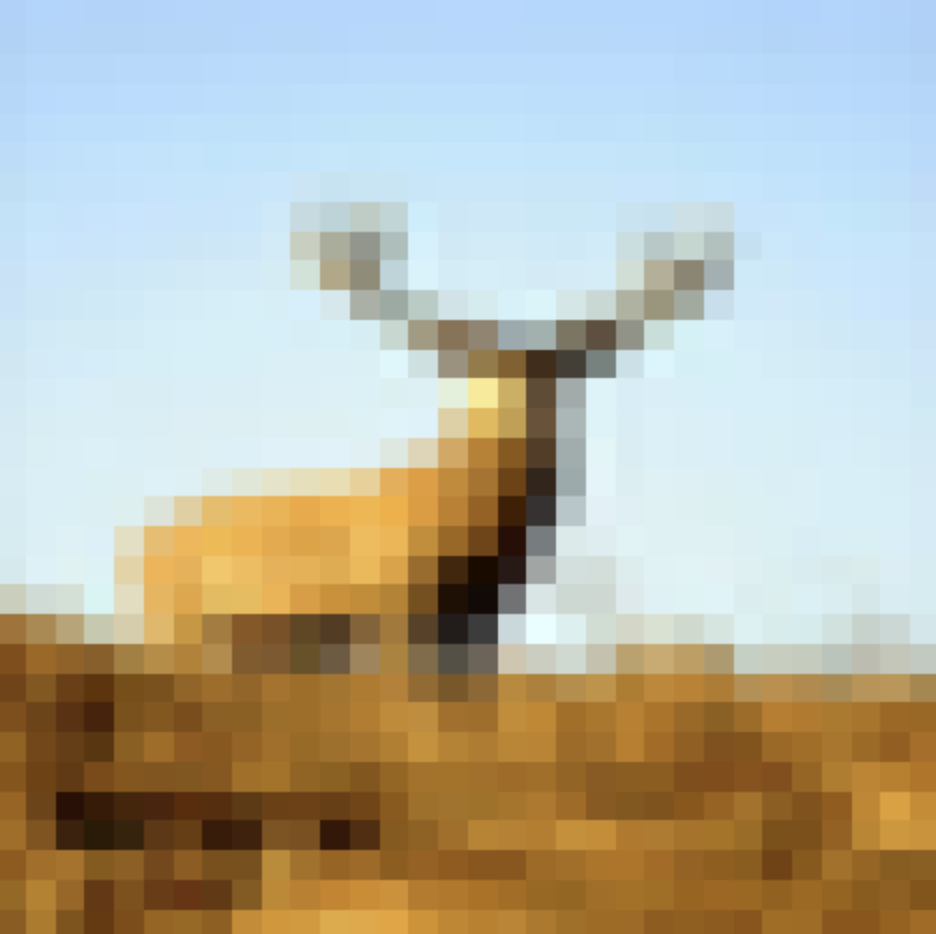}
    \caption{CIFAR10 \& CIFAR100.}
    \label{fig:second}
  \end{subfigure}
  \caption{Example images of MNIST, CIFAR10 and CIFAR100. Source: Images extracted from \cite{MNIST} and \cite{CIF}.} 
  \label{fig:graphs}
\end{figure}
CIFAR10 and CIFAR100 have the same type of colored images as seen in Figure 10.b, but the latter presents a harder challenge for machine learning as it has 100 classes and less images per class, thus the focus is set on efficient feature learning. Both these models have classes such as dogs, planes, etc. These vary in distortion, lighting and point of view. Therefore, their features are harder to learn and, at the same time, more relevant, in the sense that to detect the same pattern in a very different image of the same class, the features must be well learned.
\subsection{Model construction}
As mentioned above, we have taken inspiration from VGG CNNs to build our models. Our image sizes are small and thus the network will have less layers and perform a shallower analysis. Memtorch also offers a limitation in terms of time of computation, which is hugely affected by the size of the model to patch. For instance, for a patched model of 6 convolutions, 3 fully-connected linear layers and more than 264 hidden units, patching and inference could take several hours. Therefore, the models have to be the most efficient they can in terms of size and accuracy. 
\\ 
On top of that, image size is another aspect to consider, as the area varies with each layer according to (4). The right hyperparameters have to be chosen to avoid decreasing too much the size of the image before the features can be learned, or to not add excessive padding to the image, as it increases computation time.
Several factors impact the design of CNN architectures. Firstly, colored images pose greater challenges for pattern recognition, making datasets like MNIST, which contains simpler grayscale images, less complex by comparison. Additionally, the size of the images is crucial, as the dimensions vary across different layers. Selecting appropriate hyperparameters is essential to ensure that the image size does not reduce excessively before adequate feature extraction occurs. Moreover, avoiding excessive padding is important, as it can lead to increased computation time.
\\
The number of classes is a very relevant property of a dataset, as the model needs a better understanding of each one in order to set them apart. On top of that, more complex images need deeper CNNs with more convolutional layers to be learned.
When constructing our conventional CNNs, we adjust several parameters including the number of convolutional layers, padding, kernel size, and stride. During the training phase, we also determine the optimal learning rates and the number of hidden units. As a standard practice, all models are trained for 50 epochs.
\begin{figure}[h!]
    \centering
    \includegraphics[width=0.40\linewidth]{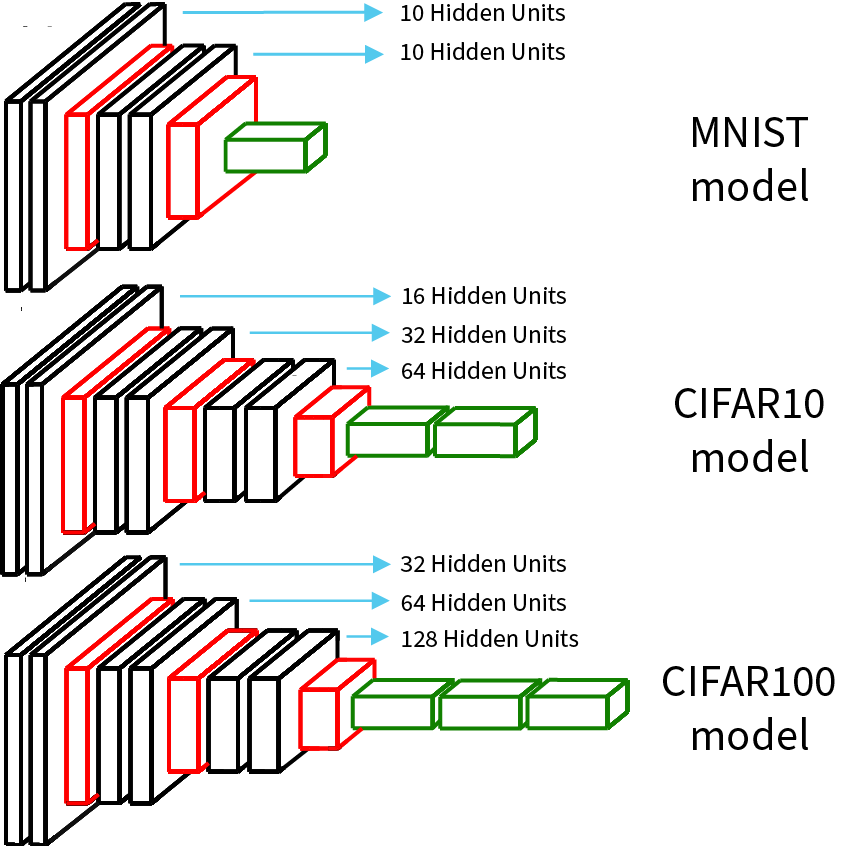}    
    \caption{Constructed models structure. The convolutional neural network (CNN) models are named after the datasets used for training. For example, the "MNIST model" refers to the CNN specifically designed and trained on the MNIST dataset. This naming convention helps clarify which model architecture is tailored to which dataset.}
    \label{fig:litio47}
\end{figure}
\subsubsection{Model Architecture Customization for MNIST, CIFAR10, and CIFAR100 Datasets}
To clarify, each model is named after the dataset used for its training, as shown in Figure 11. The so-called "MNIST" model consists of two convolutional blocks, each containing two convolution layers, batch normalization, and ReLU activation. Following each convolutional block, a pooling step is performed to simplify and reduce the image size by half. A fully-connected layer concludes the architecture, a common practice in such models, with the number of hidden units fixed at 10. The "CIFAR10" and "CIFAR100" models are structured similarly, each comprising three blocks with two convolution layers. However, "CIFAR10" features 16, 32, and 64 hidden units in each successive block, while "CIFAR100" model has twice as many in each—32, 64, and 128. To combat overfitting, dropout layers are incorporated into both models. Additionally, the fully-connected section of "CIFAR100" includes three layers, compared to "CIFAR10's" two layers. From now on, we will refer to each network by the name of its training dataset (without quotation marks).
\subsection{Training phase}
To assess the quality of the models, we implemented a K-Fold cross-validation method, dividing each dataset into 5 folds, with each fold containing designated training and validation segments. This approach allows us to evaluate how consistently the model performs across different subsets and check for any variations in convergence across the folds.
After completing the validation phase, we proceeded with training using PyTorch's designated training and testing partitions. Following this training phase, we saved the state dictionaries of the models, which include their respective weights and biases, along with their performance metrics on the test set.

\subsection{Patching model with Memtorch }
\subsubsection{Memristor model}
For this study, the chosen memristor model was VTEAM \cite{VTEAM}, which is a voltage-controlled memristor. It has a certain voltage threshold that needs to be surpassed in order for the device to display a change in its conductance state. This is a realistic implementation, as this threshold has been observed experimentally.
The VTEAM model is calibrated with parameters that replicate the behaviors of three experimentally tested memristors: a Pt-Hf-Ti memristor, a ferroelectric memristor, and a metallic nanowire memristor, thereby providing a robust foundation for simulating authentic physical phenomena.

The customizable memristor parameters are the time resolution of the simulation; $\alpha_{off}$ and $\alpha_{on}$, which control the rate of change of the state variable of the memristor at the low and high resistance state; $v_{off}$ and $v_{off}$, the thresholds from which the state variable starts to vary; $R_{off}$ and $R_{on}$, the minimum and maximum resistance states; and $k_{off}$ and $k_{on}$, which set the rate of change of the state variable when voltage is below or over a certain value. \\
As a general measure, we chose the parameters that correspond to the behavior of the Pt-Hf-Ti memristor.
\subsubsection{Crossbar Array and Mapping}
When integrating the model with the crossbar structure and mapping algorithm, several additional parameters must be considered. For example, the mapping routine provides multiple functional options. In this study, we opted for the naive mapping approach because it simplifies the translation of weights into memristor parameters. We also need to specify which types of layers are to be converted; in this case, both convolutional and linear layers are included.
Additionally, it's essential to determine the configuration of the crossbar array, whether it's a 1T1R (one transistor-one resistor) or a simpler 1R (one resistor) setup. This distinction is made using a boolean variable. Transistors are preferred in our model because they help to accurately map weights by minimizing interference from adjacent memristors, enhancing the precision of weight adjustments and deletions. Consequently, this setting has been enabled (set to true).

Memtorch allows us to simulate the programming routine, which sets how weights are mapped with voltage pulses. It has been set to none in order to prevent going through excessive steps in the simulation. If our crossbar array is 1R, a programming routine should be specified, as mapping weights requires specific instructions. To block out and select memristors is much more complex without transistors. \\
Additionally, it is very relevant to modify tile shapes to adjust to each CNN. Memtorch API recommends tile shapes of $128\times128$, $256\times64$ or $256\times256$. Nevertheless, smaller arrays of $64\times64$ are also an option.
\\
In order to keep reasonable physical variables, a maximum input voltage is specified to prevent simulating unrealistic voltages that would damage the device. The threshold was adjusted to study its effect on weight mapping. \\
Shifting our focus to the external parts of the circuitry, the ADC has to be given a resolution of bits that are used to interpret the analog input and an overflow percentage. When an input is higher than the ADC's limit, it overflows, giving the analog input the highest possible digital output. The percentage of overflowing inputs was set to 0.\\
When converting weights to conductances, a quantization method is required. Our models were patched using linear quantization, which effectively maps the weights into a discrete number of conductance states evenly. Quantization can be done logarithmically as well as exponentially.\\ 
Finally, we can customize parameters like the mapping scheme, which determines how weights are assigned to memristors. We've opted for the double column scheme, which utilizes two memristors per weight to accommodate negative values. While a single column scheme is an alternative, it does not support negative values, which could compromise the accuracy of weight mapping and subsequent calculations.
\subsection{Experiments}
Using the available non idealities offered by Memtorch, we will conduct a series of inference phases with different combinations.
\begin{itemize}
    \item Effect of varying max input voltage on accuracy.
    \item Different tiles sizes of $64\times64$, $128\times128$ and $256\times256$ with varying ADC resolutions from 2 to 10 bits.
    \item Stochastic effects varying $\sigma$ in relation to conduction states. 
    \item Device faults for stuck-at HRS and LRS.
    \item Endurance and retention effects.
\end{itemize}
CIFAR100 will be only used for the first experiment. For the rest, only MNIST and CIFAR10 will be patched as they are computationally lighter and easier to work with.
\section{Results}
\subsection{Training of CNNs}
Following our methodology, we will first comment on the quality of our models. To study their behavior on the datasets, K-Fold cross validation was performed. Hereunder the results are presented for each CNN in Figure 12.
\begin{figure}[ht]
  \centering
  \begin{subfigure}[b]{0.49\textwidth}
    \centering
    \includegraphics[width=\textwidth]{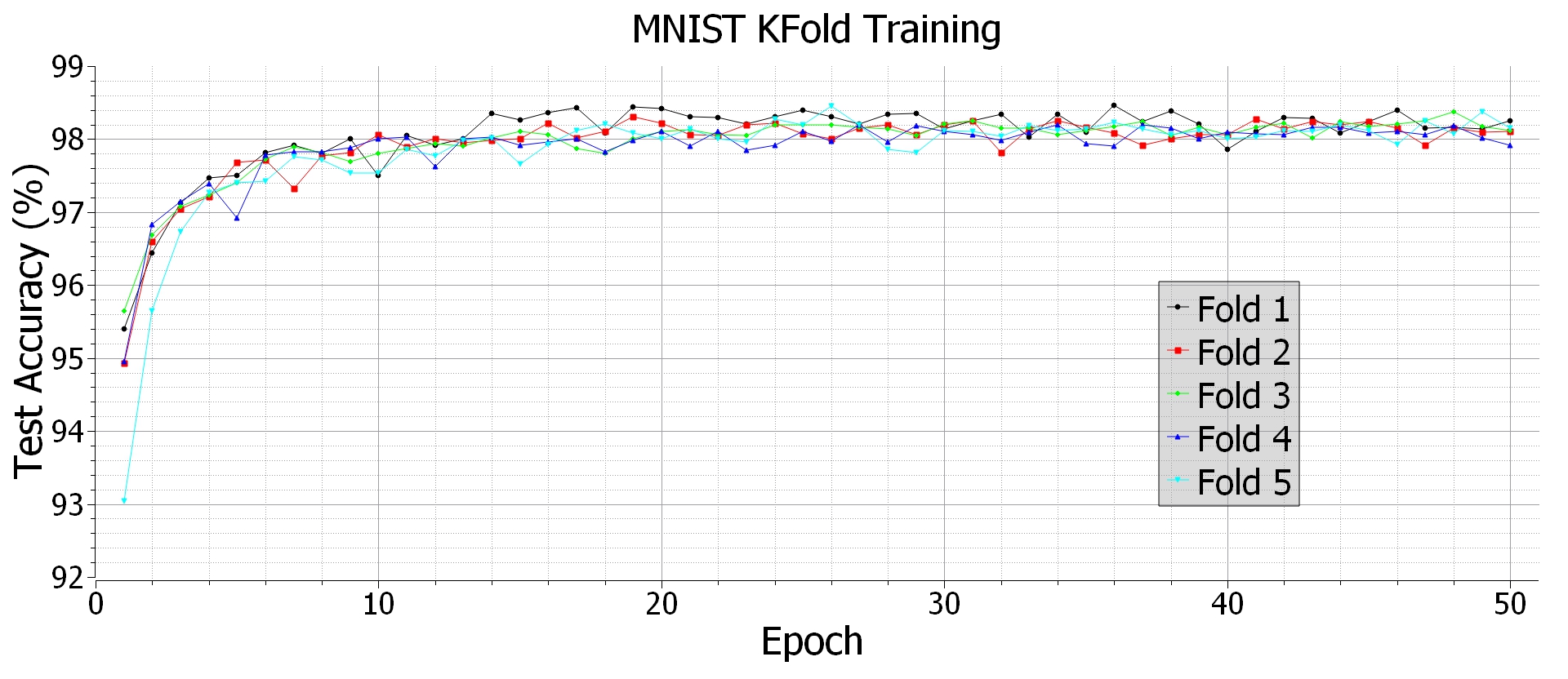}
    \caption{}
    \label{fig:first}
  \end{subfigure}
  \hspace{0.01\textwidth}
  \begin{subfigure}[b]{0.49\textwidth}
    \centering
    \includegraphics[width=\textwidth]{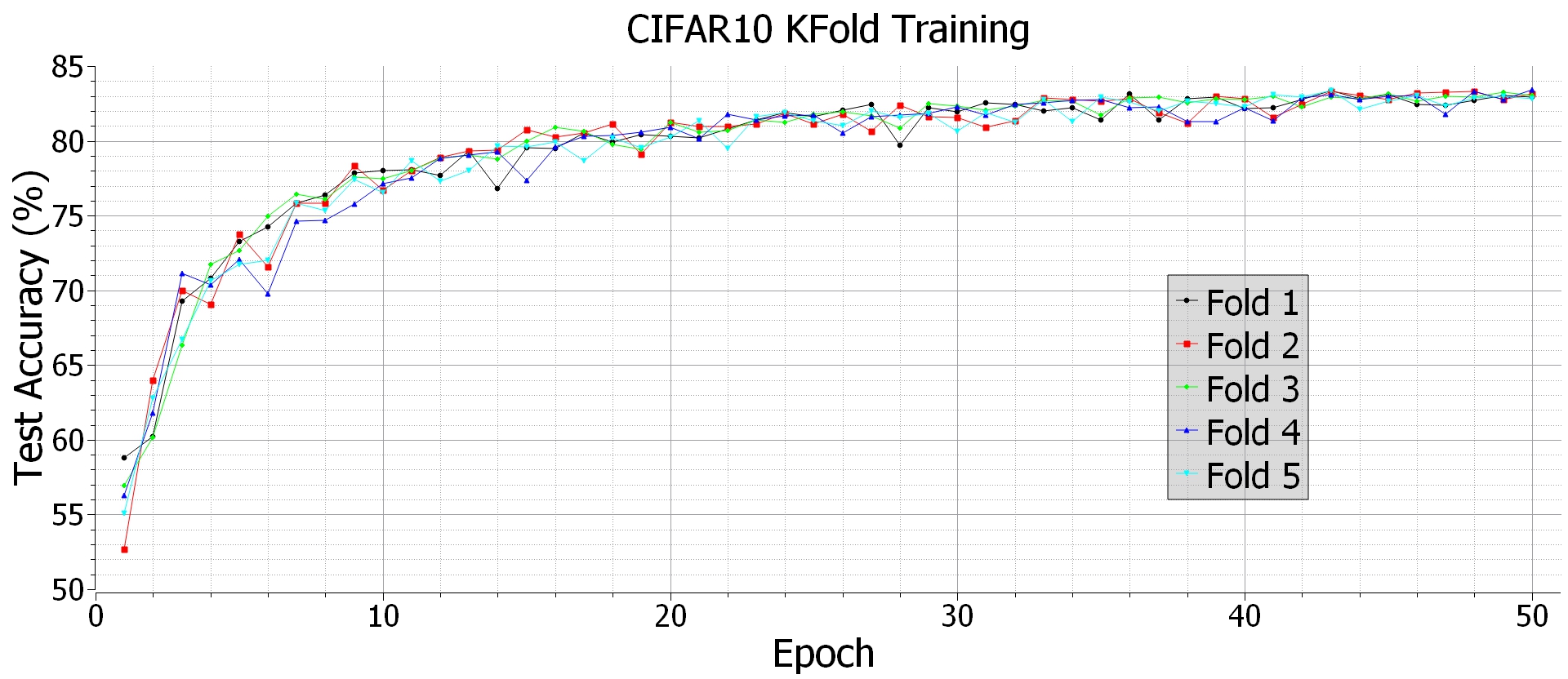}
    \caption{}
    \label{fig:second}
  \end{subfigure}
  \begin{subfigure}[b]{0.49\textwidth}
    \centering
    \includegraphics[width=\textwidth]{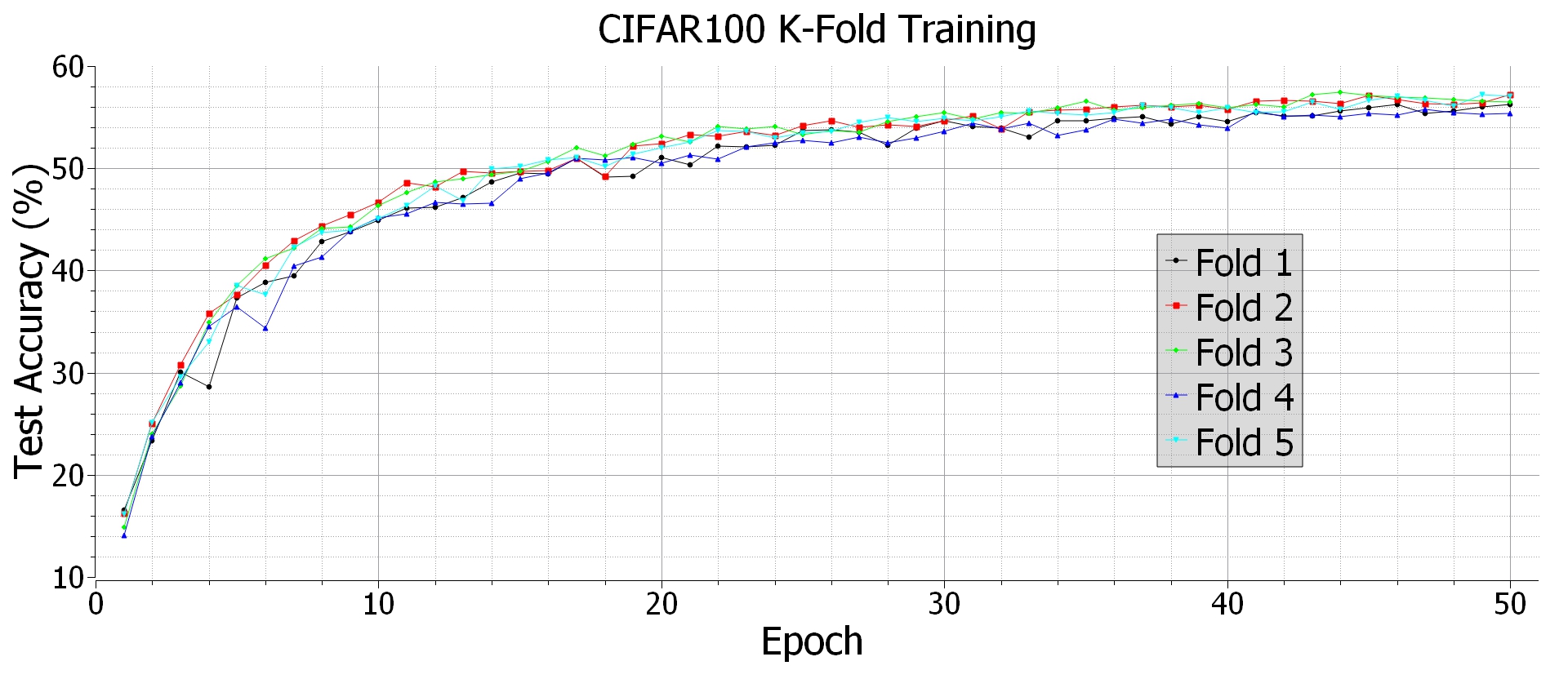}
    \caption{}
    \label{fig:second}
  \end{subfigure}
  \caption{Cross validation accuracy results for MNIST, CIFAR10 and CIFAR100.}
  \label{fig:graphs}
\end{figure}    
\\ \\
As seen in the graphs, the training of each fold behaved similarly, which showed proof that the models do not act differently in the training phase depending on the slice of the dataset. Cross validation also allowed us to measure the quality of the model in terms of the reached accuracy over the test set, which we obtained by averaging the obtained accuracies of the folds, displayed on Table 2.a. \\
For MNIST, the convergence happens rather quickly and it has a very high starting accuracy. This shows the simplicity of the dataset, which is quickly learned by the model. Both CIFAR10 and CIFAR100 take more epochs to converge into a final value.
\begin{table}[h!]
\centering
\begin{subtable}[t]{0.45\textwidth}
    \centering
    \begin{tabular}{|l|c|l|l|}
    \hline
    \rowcolor[HTML]{C0C0C0} 
    \cellcolor[HTML]{C0C0C0}K-Fold &  \cellcolor[HTML]{9B9B9B}MNIST & \cellcolor[HTML]{9B9B9B}CIFAR10 & \cellcolor[HTML]{9B9B9B}CIFAR100 \\ \hline Acc. (\%)
    & 98.35 & 83.34 & 56.75 \\ \hline
    \end{tabular}
    \caption{}
\end{subtable}
\hfill
\begin{subtable}[t]{0.45\textwidth}
    \centering
    \begin{tabular}{|l|c|l|l|}
    \hline
    \rowcolor[HTML]{C0C0C0} 
    \cellcolor[HTML]{C0C0C0}Test & \cellcolor[HTML]{9B9B9B}MNIST & \cellcolor[HTML]{9B9B9B}CIFAR10 & \cellcolor[HTML]{9B9B9B}CIFAR100 \\ \hline Acc. (\%)
    & 98.73 & 83.34 & 58.85 \\ \hline
    \end{tabular}
    \caption{}
\end{subtable}
\caption{Test accuracies for K-Fold training (a) and PyTorch partition training (b).}
\end{table} \\
Yet, after this characterization of the quality of the CNN, another training phase is conducted using the partitions given by PyTorch for each dataset. We have saved the state dictionary of the epoch for which the model had a higher test accuracy over the test set. This decision was made in order to later patch the most accurate model possible. The models' final accuracy after training over the test set can be seen in Table 2.b.
\subsection{Memristive inference experiments}
In this section, the effects in inference accuracy of the patching of the CNNs using Memtorch will be discussed. Testing phases varied on time depending on the model and parameters. MNIST inference took 196 seconds with tile size of $64\times64$, while CIFAR10's took 680 seconds and CIFAR100's 1300 seconds using the same settings. \\
The Memtorch simulations were computed with a NVIDIA GTX 1060 GPU, Intel Core i7-8750H CPU with 6 cores and 16GB of RAM on a Python 3.10.12 environment. The operating system was Ubuntu 22.04 using Windows Subsystem for Linux. Nevertheless, the training of the digital models was carried out in Google Colab using an A100 GPU. All models averaged epochs of 15 seconds.
\subsubsection{Effect of max input voltage}
The maximum input voltage (Max Input) was adjusted to observe its impact on the accuracy of our models. As depicted in Figure 13, the accuracy across the test set stabilizes for voltages above 6V for all models.
Nevertheless, MNIST has a rather high initial precision for 1V, whereas CIFAR10 deteriorates to a 14\% and CIFAR100 to a 1\%. 
\begin{figure}[h!]
  \centering
  \includegraphics[width=0.70\textwidth]{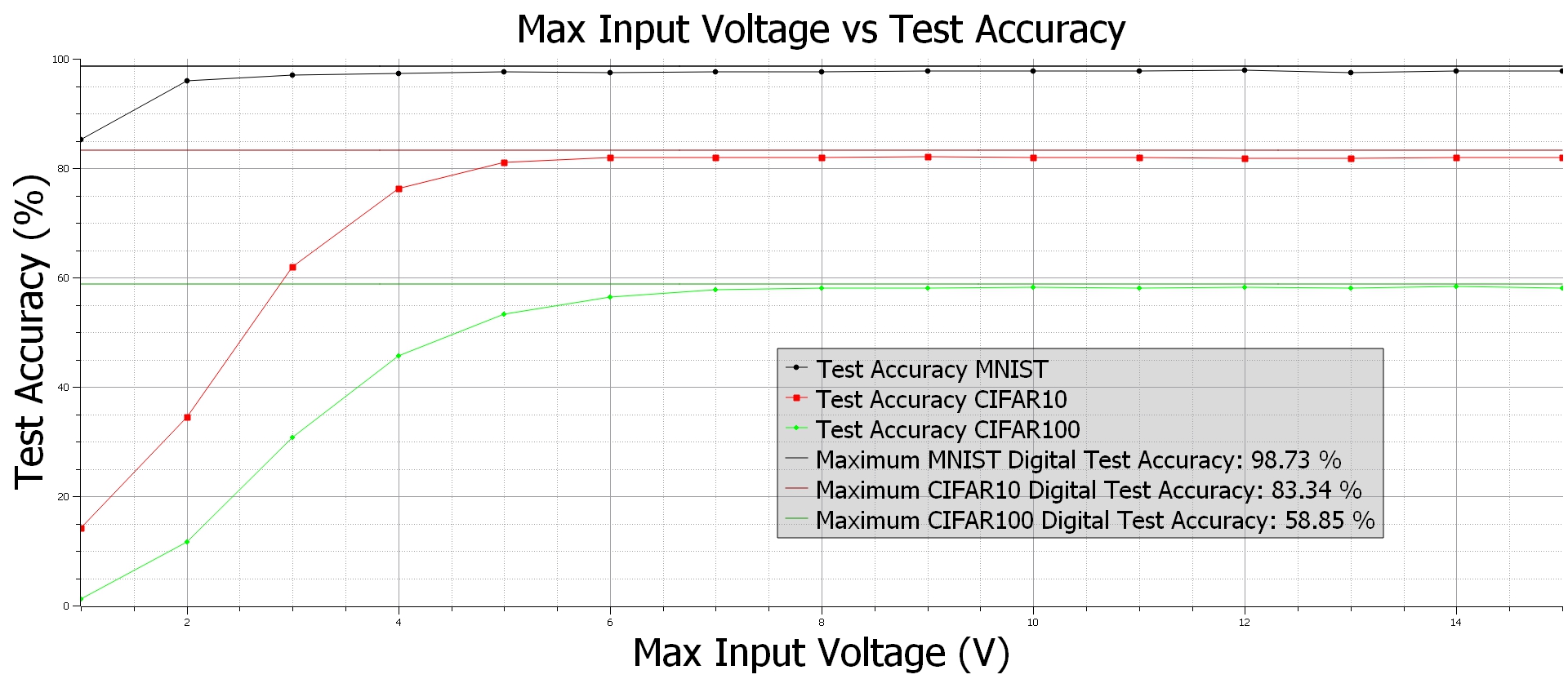}
  \caption{Impact of variable maximum input voltage on test set accuracy during inference for MNIST, CIFAR10, and CIFAR100.}
  \label{fig:graphs}
\end{figure} 
\subsubsection{ADC resolution vs Tile Size}
The model was patched for each combination of hyperparameters, i. e., an ADC resolution going from 2 to 10 bits in intervals of two and tile shapes of $64\times64$, $128\times128$, $256\times256$. 
\begin{figure}[]
    \centering
    \begin{subfigure}[b]{0.48\textwidth}
    \centering
    \includegraphics[width=\textwidth]{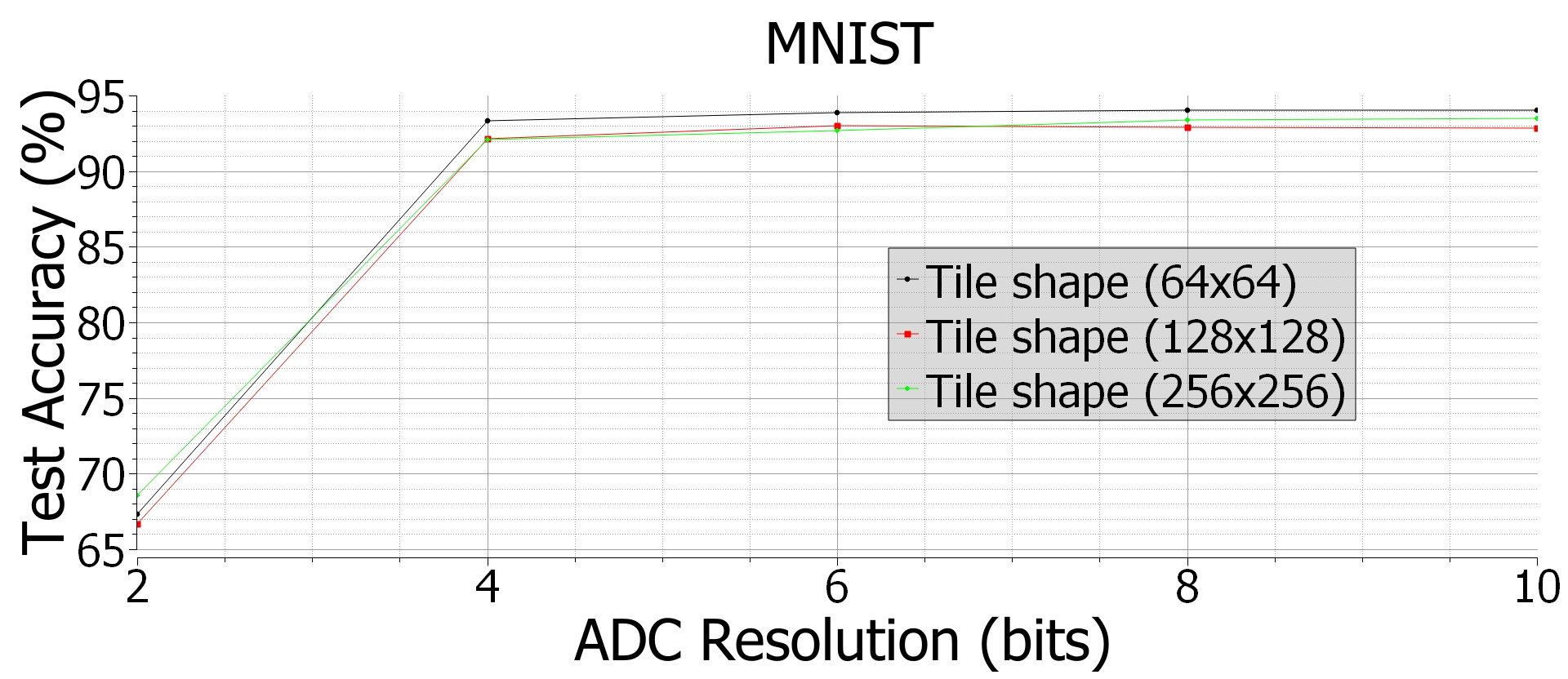}
    \caption{}
    \label{fig:first}
  \end{subfigure}
  \begin{subfigure}[b]{0.50\textwidth}
    \centering
    \includegraphics[width=\textwidth]{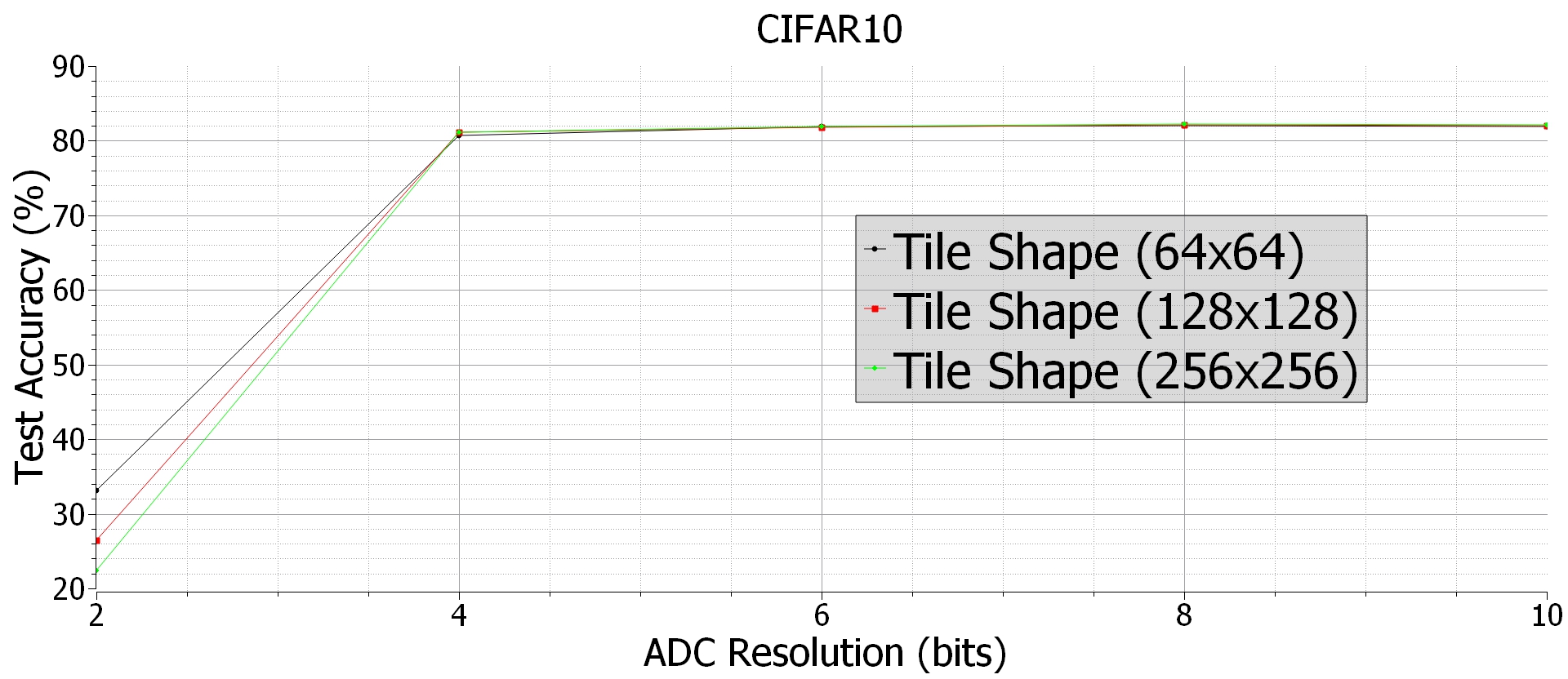}
    \caption{}
    \label{fig:first}
  \end{subfigure}
  \caption{Test accuracy variation for different ADC resolutions and varying tile shapes for (a) MNIST and (b) CIFAR10.}
\end{figure} 
As seen in Figure 14, the smaller tile shape offers slightly better results consistently for all ADC resolutions for CIFAR10. This fact proves the necessity of choosing an appropriate crossbar array size for effective performance.\\
\subsubsection{Stochastic Effects vs Conduction States}
Random effects can be modeled by means of a sigma value that scales the stable resistance states of the memristor, representing the inherent variability of the device. This causes a loss in accuracy as mapped weights are altered. On top of these stochastic effects, we can apply different numbers of conductive states to observe which non-ideal behavior is more dominant and their interrelation.
\begin{figure}[h!]
  \centering
  \includegraphics[width=\textwidth]{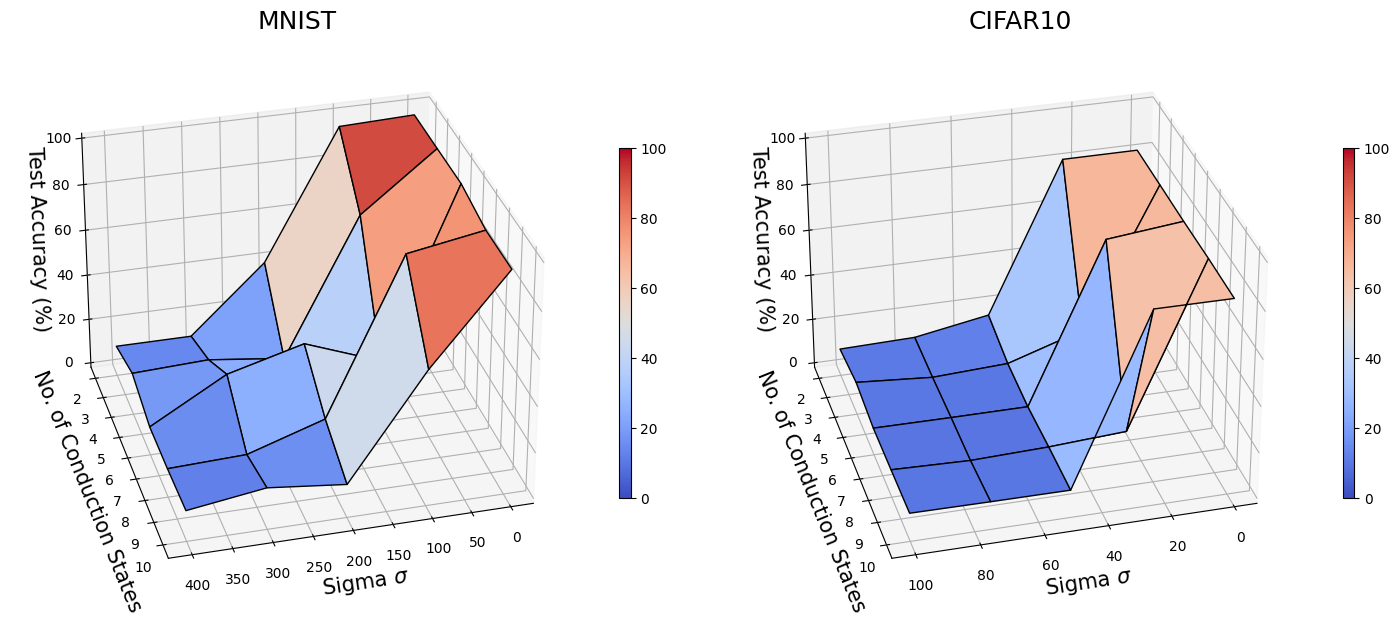}
  \caption{Stochastic effects vs Number of Conduction States for MNIST and CIFAR10.}
\end{figure}\\
In Figure 15, we see the deterioration caused by increasing the variance, whereas the conductance states do not affect accuracy significantly.
\subsubsection{Device Faults}
Memristors are susceptible to getting stuck into their high and low resistance states, rendering them unusable. This can be modeled by introducing a probability of falling into these states. A 3D plot in Figure 16 was constructed for both MNIST and CIFAR10 to observe how these probabilities for HRS and LRS relate.
\begin{figure}[h!]
\centering
\includegraphics[width=\textwidth]{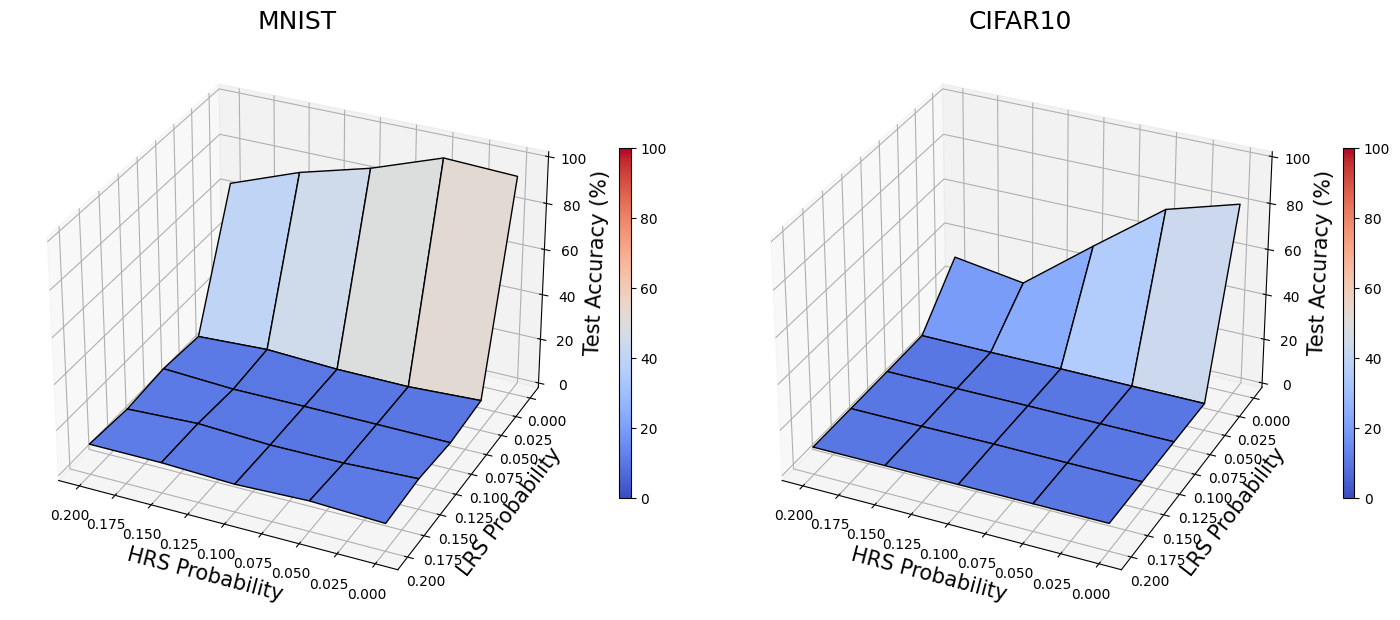}
\caption{Device faults effect on test accuracy for MNIST and CIFAR10.}
\end{figure} 
LRS probability is clearly dominant for the same range of parameters as HRS. For a null probability of getting stuck at a LRS, HRS probability deteriorates the accuracy of the model less abruptly. Nevertheless, a 5\% chance of stuck-at LRS effectively reduces accuracy to minimum. 
\subsubsection{Endurance and Retention Effects}
Memristors deteriorate with cycles of set/reset operations and their conductance state slowly drifts. These non-idealities can be simulated using the retention and endurance Memtorch submodules. Then, for each 10-class model two simulations have been carried out obtaining the results displayed in Table 3.
\begin{table}[h!]
\centering
\begin{tabular}{c|c|c|}
\cline{2-3}
 & \cellcolor[HTML]{C0C0C0}MNIST & \cellcolor[HTML]{C0C0C0}CIFAR10 \\ \hline
\multicolumn{1}{|c|}{\cellcolor[HTML]{C0C0C0}Endurance} & 52.47 \%                      &     36.01 \%                          \\ \hline
\multicolumn{1}{|c|}{\cellcolor[HTML]{C0C0C0}Retention} & 93.94 \%                      &     81.89 \%                            \\ \hline
\end{tabular}
\caption{Test accuracies for the endurance and retention experiments.}
\end{table}
For retention, drift must be specified as it accounts for the shift in conductance with time. In our experiment we simulated 
conductance state drift of 0.1 per second over a period of 10,000 seconds.
Next, to evaluate endurance, we modeled the decrease in accuracy resulting from 10,000 set/reset cycles. We also considered the temperature, as this parameter significantly influences deterioration over time, setting it at T=350 K.
\section{Analysis of Results} 
\subsection{PyTorch Models}
The three models used varied in shape, depth and amount of layers. This was beneficial to test the simulator for a range of inputs. MNIST offered great results due to the easy-to-learn patterns of the inputs, reaching a 98.73 \% accuracy over the test set. On the other hand, the CIFAR10 dataset required a deeper network in order to show an acceptable accuracy of 83.34\%. Finally, CIFAR100's dataset converged to 58.85\%. This value is not satisfactory for computer vision tasks, nevertheless, the computational limitations of the GPU when patching models with Memtorch made impossible the improvement of the model. Improvements would involve more convolution layers, more hidden units or applying more advanced techniques. Residual layers or LeakyReLU are examples of beneficial additions for attaining higher accuracy rates.
\subsection{Maximum Input Voltage} 
Observing the results in Figure 13, it can be inferred that  Maximum Input Voltage (MIV) effectively alters inference. Excessively low voltages can lead to severe degradation of accuracy. Further, excessively high voltages can damage memristors due to thermal effects and electromigration, consequently decreasing performance and longevity.\\
Thus, a reasonable value has to be achieved in order to map weights and perform MVMs properly. For the proposed models, it was seen how voltages over 6V did not improve accuracy as the value was constant from that point. Nonetheless, the maximum accuracy after patching is evidently the maximum accuracy reached after training. This value was almost achieved, as the models fell approximately 1\%  the maximum value as seen in Table 4. 
\begin{table}[h!]
\centering
\begin{tabular}{c|c|c|l|}
\cline{2-4}
\rowcolor[HTML]{C0C0C0} 
\cellcolor[HTML]{FFFFFF}                                   & {\color[HTML]{000000} MNIST} & {\color[HTML]{000000} CIFAR10} & {\color[HTML]{000000} CIFAR100} \\ \hline
\rowcolor[HTML]{FFFFFF} 
\multicolumn{1}{|c|}{\cellcolor[HTML]{C0C0C0}Digital (\%)} & 98.73                        & 83.34                          & 58.85                           \\ \hline
\rowcolor[HTML]{FFFFFF} 
\multicolumn{1}{|c|}{\cellcolor[HTML]{C0C0C0}MIV (\%)}  & 97.85                        & 82.12                          & 58.32                           \\ \hline
\end{tabular}
\caption{Maximum test accuracy reached in the range of MIV proposed for each model.}
\end{table}
For the remainder of the simulations, a voltage of 9V was maintained to ensure satisfactory accuracy for analyzing other parameters. Yet, in a realistic memristor 9V is a rather high voltage input. High enough voltages could damage the crossbar array, increasing temperature and consuming more energy than intended. \\
Max input voltage directly controls how weights are mapped and how input images enter through the DAC converter. Lower values may not be able to map bigger weights and thus this threshold can overflow the mapping if a high enough value is not selected. In MNIST's case, the accuracy was high from 1V, meaning than the model's performance is not as dependent on precise weight mapping.
\subsection{Tile shape vs ADC Resolution}
The ADC resolution resulted to a be a bottleneck for accuracy for resolutions of less than 6 bits as seen in Figure 14. Nevertheless, over that value the accuracies did not vary noticeably. This is due to the fact that ADCs must convert the analog signal into a discrete digital one. Thus, the more bits of resolution the more accurately the device will be able to represent changes in the analog signal. Insufficient resolution may cause poorly translated output signals and consequently inaccurate labels in image classification tasks. On the other hand, ADC resolution must be kept the lowest possible as it has a bigger impact on energy consumption the higher it is \cite{ADC}. Therefore, 8 bits was used as a standardized value for ADC resolution to secure no impact on accuracy.
\\
Depending on the tile size of the memristor crossbar arrays, the best fit algorithm mapping the network may divide the weight matrix into several arrays \cite{sims}. Although it is a more complex process it has advantages. Errors are localized, whereas with bigger tiles an error greatly decreases accuracy. In addition, smaller tiles allow for parallel processing. Small tiles come with disadvantages too as they make the design much more complex. This is due to the fact that tiles much be interconnected and be synchronized. Thus, there is an inherent latency between tiles. In conclusion, a middle ground between over separating the matrix and mapping it as a whole must be reached via testing. As seen during simulations, the sizes used did not differentiate noticeably in terms of accuracy. Nevertheless, the smallest tile shape, $64\times64$, was very useful throughout the work as it allowed for much faster inference than the other two. 
\begin{table}[h!]
\centering
\begin{tabular}{|c|c|c|c|}
\hline
\rowcolor[HTML]{C0C0C0} 
\cellcolor[HTML]{EFEFEF}MNIST    & $64\times64$ & $128\times128$ & $256\times256$ \\ \hline
\cellcolor[HTML]{C0C0C0}Time (s) & 196          & 464            & 1497           \\ \hline
\end{tabular}
\caption{Inference time for different tile shapes for MNIST model.}
\end{table}
A tendency  can be observed in Table 5 where smaller tiles meant faster testing for MNIST, which is consistent with the parallelism that they allow while computing. This increase in speed was seen for all models.
\subsection{Stochastic Effects and Number of Conductance States}
Setting a finite number of conductance states did not alter accuracy as it can be seen by observing the values for $\sigma$=0 in Figure 15. Therefore, the calculated accuracy for each conductance state can be used as an extra iteration of the simulation to observe how a certain value of $\sigma$ has a range of possible accuracies. This variability that is introduced models how devices display random behaviors based on material defects, thermal effects, etc. As a result, the $R_{off}$ and $R_{on}$ are affected by this variance, which degrades performance. In fact, they can overlap if the $\sigma$ is big enough, effectively sharing a conductance state \cite{memT2}.
\\ 
Circling back to the number of conductance states, these discretize the values of the mapped weights. For both our models two conductance steps were sufficient to quantize weights. This can be interpreted as a result of using Memtorch on a simple model that does not need more resolution to map the weights. A more complex network may be more susceptible to the number of conductance states available to map their weights, as a binary representation lacks the necessary range for precise mapping. Yet, binary devices are easier to manufacture, since multi-level memristors pose a harder challenge to produce on a large scale \cite{Cond}.   \\
The stochastic effects were clearly seen. Higher $\sigma$ resulted in more randomness in the results. For $\sigma$=100 in MNIST it was seen how the standard deviation was still low enough that the accuracies were not affected as drastically.
\subsection{Device Faults}
Memristors undergo a process of electroforming to operate at a pristine state. If this practice is not performed well, it can lead to memristors that do not operate properly \cite{saf}. For instance by getting fixed in a high resistance state if conductive filaments are not formed. Memtorch accounts for this as well as for LRS stuck-at faults (SAFs) by means of assigning probabilities to these faults.\\
As it was seen in Figure 16, LRS probability lowered accuracy much more than HRS. For instance, in Table 6, a 5\% probability for LRS SAF meant a decrease to the minimum accuracy.
\begin{table}[h!]
\centering
\begin{tabular}{|
>{\columncolor[HTML]{C0C0C0}}c |c|c|}
\hline
\cellcolor[HTML]{EFEFEF}SAF CIFAR10 & \cellcolor[HTML]{C0C0C0}LRS 0\% & \cellcolor[HTML]{C0C0C0}LRS 5\% \\ \hline
HRS 0\%                             & 81.95 \%                        & 10.01 \%                        \\ \hline
HRS 5\%                             & 72.88 \%                        & 10.04 \%                        \\ \hline
\end{tabular}
\caption{Accuracy Tests for Various SAF Probabilities in CIFAR-10}
\end{table}
This can be explained by means of interpreting what these states represent. A high resistance state stores a small weight as conductance is inversely proportional to resistance. 
Essentially, smaller weights have a lesser impact on the network's output compared to larger weights, which have the potential to activate additional neurons more effectively.
\subsection{Retention and Endurance}
Devices present a limited number of cycles as their characteristics start to degrade with use. Memtorch allowed us to account for the drift of the stable high and low resistance values after a number of cycles. The number of set/reset cycles was added to the inputs, in our case, 10,000 cycles were sufficient to observe a degradation in performance. During set/reset cycles, the conductive filaments of the device are created and broken \cite{endret}. This process degrades the $R_{on}/R_{off}$ ratios, effectively decreasing the dynamic range of the memristor to map weights. As seen in the experiment, endurance chunked accuracy to 52.47\% in MNIST and 36.01\% in CIFAR10. 
Another element that decreases endurance is temperature, which can damage the device. High temperatures can alter conductance acting as a bias \cite{temp}. On top of that, they are responsible for material degradation. \\
As we know, a memristor is a highly relevant device for in-memory computing, since it is able to maintain its state. Yet, its conductance does vary with time slightly. Memtorch grants us a framework to simulate this drift by specifying total time and rate of change of conductance. For 10,000 seconds and a conductance drift of 0.1, accuracy did not decrease noticeably for CIFAR10 as it was kept at 81.89\%. In MNIST, accuracy was 3\% lower than that of an ideal memristive network.\\
Overall, it can be seen how retention effects are less prominent than endurance non-idealities as many set/reset cycles may take place in under a second \cite{quant}. Thus endurance decreases performance quicker than retention effects.

\section{Conclusion}
This research advances the simulation of neuromorphic systems utilizing analog devices by presenting a methodology designed to assess nonlinear behaviors in CNNs using memristive devices. It involved designing three distinct CNN architectures, each named after and tailored to accommodate the varying complexities of their respective datasets: MNIST, CIFAR10, and CIFAR100.

Five experiments were conducted using Memtorch as a simulation framework. Firstly, for ideal conditions the effect of varying MIV, tile shape and ADC resolution was tested to tune hyperparameters for later experiments. A MIV of 9V, a tile shape of $64\times64$ and 8 bits of resolution offered a loss of less than 1\% in test accuracy with respect to its digital counterpart. Additionally, this set of parameters optimizes acceleration while keeping the precision loss bounded.

Continuing with non-ideal behavior, a varying number of conductance states in relation to stochastic effects resulted in virtually no loss of accuracy for the range of conductance states. On the other hand, randomness deteriorated performance drastically, especially for CIFAR10, which was more susceptible than MNIST to inherent device variability. Device faults showed the landslide in accuracy loss between LRS and HRS SAFs, showing evidence of how bigger weights are more influential to outputs. Finally, endurance and retention were simulated to test the evolution in performance of memristive devices. Cycle to cycle degradation was dominant over retention, which caused less than 3\% loss in accuracy.
\\
In conclusion, following the proposed methodology, near-digital accuracies were reached for the proposed models using Memtorch. The CNNs constructed were built with a set of hyperparameters that optimized accuracy and size. Then models were patched using a set of realistic physical parameters to analyze their behavior and assess the quality of the simulator for PIM with ReRAM devices. The results were satisfactory, strengthening the importance of analog computing in today's computational landscape.

\end{document}